\documentclass[twocolumn]{article}

\usepackage[pagenumbers]{arxiv}

\usepackage[a4paper, margin=20mm]{geometry}

\usepackage[linesnumbered,ruled,vlined]{algorithm2e}
\usepackage{graphicx}
\usepackage{amsmath}
\usepackage{amssymb}
\usepackage{booktabs}
\usepackage{hyphenat}
\usepackage[inline]{enumitem}
\usepackage[breaklinks,colorlinks]{hyperref}
\usepackage{multirow}

\SetKwComment{tcp}{\% }{}
\SetKwInOut{Input}{Input}
\SetKwInOut{Output}{Output}

\usepackage[capitalize]{cleveref}

\newcommand{\best}[1]{\textbf{\textcolor{red}{#1}}}

\begin{document}

\title{Multibranch Ensemble Head in Vision Foundation Models for Efficient Cucumber Phenological Stage Classification}
\author{Kostiantyn Khabarlak \\
  Dnipro University of Technology, Ukraine\\
  {\tt\small Khabarlak.K.S@nmu.one}
}

\maketitle

\begin{abstract}
Vision Foundation Models (VFMs) pretrained with self-supervised learning, such as DINOv2, provide highly transferable features, yet adapting them to small, domain-shifted datasets is still challenging. We propose a regularizing multibranch ensemble head: a training and inference scheme that pairs low- and high-capacity branches, samples a single branch stochastically at each training step, and averages both branches in probability space at inference. The design adds negligible parameters and essentially no inference overhead over a single heavy head, while improving the accuracy-efficiency trade-off under data-scarce conditions. Accurate classification of plant growth stages is critical for precision agriculture, enabling timely decisions on fertilization, irrigation, and harvesting. Therefore, as a challenging practical case we use cucumber phenological stage classification, in which flowers and fruits are frequently occluded by foliage, and stage transitions manifest primarily as subtle changes in leaf morphology. The NALEF Food Safety dataset contains images of five cucumber growth stages captured under diverse real-world conditions. On it we evaluate a broad range of backbones, including DINOv2 and DINOv3 ViT and ConvNeXt variants, against strongly supervised counterparts (ResNet, EfficientNet, ViT). The proposed DINOv2 ViT-S with multibranch head achieves a Macro F1 of 0.978 and Cohen's Kappa of 0.971, matching the performance of models more than ten times larger in parameters and GFlops, while requiring only 12.2~GFlops and 22.3M parameters, making it deployable on smartphones and edge devices for phenological stage classification.
\end{abstract}


\blfootnote{
  This article has been published in the peer-reviewed conference proceedings. Citation: \nohyphens{K. Khabarlak, ``Multibranch ensemble head in vision foundation models for efficient cucumber phenological stage classification,'' in Proceedings of the International Scientific and Practical Workshop ``Advanced Digital Technologies: Intelligent Systems, Modeling, and Data-Driven Control'' (ADT-ISMDDC 2026) Co-Located with ICST 2026, in CEUR Workshop Proceedings, vol. 4300. Odesa, Ukraine: CEUR-WS.org, 2026, pp. 251--262. doi: \href{https://doi.org/10.83095/ceurws_04300_018}
  {10.83095/ceurws\_04300\_018}}.

  \href{https://www.youtube.com/watch?v=CxQJAUJ6NSQ}%
  {Video presentation available on YouTube.}
}

\section{Introduction}

The timing and progression of plant developmental phases are highly dependent on environmental factors, particularly temperature, precipitation, and photoperiod~\cite{PlantPhenologicalResearch2022}. Accurate classification of plant growth stages is critical for optimizing agricultural management, including decisions on fertilization, herbicide application, and harvesting. Automating this process reduces reliance on labor-intensive and subjective manual analysis, ultimately supporting higher and more consistent crop yields.

Monitoring of plant growth can be performed using ground-level cameras, unmanned aerial vehicles (UAVs), or remote-sensing satellite imagery.

Remote-sensing satellite imagery can be used to estimate growth stages of large farmlands. For instance, Sentinel-2 satellites provide public access to multi-spectral imagery, including red-edge bands specifically designed for vegetation analysis~\cite{Sentinel2}. However, this approach is unsuitable for use in cloudy conditions or in greenhouses, and its low spatial resolution leads to poor performance at early growth stages when the canopy does not fully cover the field, or for small fields.

UAVs and stationary cameras offer much greater flexibility and precision. They can be used regardless of weather conditions and inside greenhouses. Additionally, smartphones enable rapid, on-site plant growth evaluation by non-experts. Camera-based plant growth monitoring is often formulated as a classification task~\cite{EfficientNetStrawBerryGrowthClassification,ResNet50ButterheadLettureGrowthStage}.

In particular, automated classification of cucumber growth stages from visual data remains an open problem despite the crop's considerable agronomic significance. While computer vision methods have been applied to cucumbers across several related tasks including disease detection~\cite{CucumberDiseaseDetection}, fruit localization for robotic harvesting~\cite{YoloCucumberPlantDetection}, quality grading~\cite{CucumberFruitClassificationCNN}, and morphological phenotyping~\cite{CucumberFruitKeypointForPhenotyping}, the problem of growth stage classification has received comparatively little attention in the literature. This task poses unique challenges: flowers and fruits are frequently occluded by foliage, and cucumber fruit shares near-identical coloration with the surrounding canopy. As a result, transitions between intermediate stages manifest primarily as subtle changes in leaf morphology.

Vision-based plant monitoring is typically performed by fine-tuning a backbone trained with strong supervision on a large labeled dataset. However, recent works have shown that backbones trained with self-supervised objectives offer far more universal features that can be adapted to diverse downstream tasks using a simple linear probe~\cite{DINOv2,DINOv3}. These Vision Foundation Models (VFMs) have demonstrated applicability across object classification, detection, tracking, scene depth estimation, and many other tasks; yet their use for plant growth stage classification remains comparatively underexplored.

In this work we study how to adapt self-supervised VFMs to small, domain-shifted image datasets through the design of the classification head. We introduce a \emph{regularizing multibranch ensemble head}: a general training and inference scheme in which a low-capacity linear branch and a higher-capacity MLP branch are attached to the backbone. A single branch is sampled stochastically at each training step, and the two branches are averaged in probability space at inference. This yields an in-model ensemble that regularizes head training under data scarcity while adding negligible parameters and essentially no inference cost. We use cucumber phenological stage classification as a challenging fine-grained case study: on the recently introduced NALEF Food Safety~\cite{NalefPlantGrowth} dataset, the proposed head substantially improves the accuracy-efficiency trade-off over the DINOv2 ViT-L fine-tuning approach previously reported for this dataset~\cite{KhabarlakNalefChallenge2026}.

Our main contributions are as follows:
\begin{enumerate}
  \item We propose the regularizing multibranch ensemble head, a lightweight, backbone-agnostic scheme for training classification heads on top of Vision Models under data-scarce conditions, combining stochastic per-step branch sampling with probability-space aggregation at inference.
  \item Using a cucumber phenological stage classification dataset, we systematically compare a broad range of self-supervised (DINOv2, DINOv3) and strongly supervised backbones, and show that self-supervised pretraining is particularly beneficial in the small-model regime despite the large domain gap between its pre-training data and plant imagery.
  \item The resulting DINOv2 ViT-S model substantially improves over existing solutions on the NALEF Food Safety dataset at a small fraction of its computational cost, requiring few GFlops and parameters and remaining deployable on smartphones and edge devices.
\end{enumerate}

\section{Literature Overview}

Fine-tuning from a strongly pretrained backbone is the standard approach for training deep neural networks on downstream tasks such as plant growth stage classification. The authors of~\cite{EfficientNetStrawBerryGrowthClassification} collect a strawberry growth dataset and report that a fine-tuned EfficientNet-B7 achieves 83.7\% classification accuracy. In~\cite{BasilYoloGrowthStage}, the authors highlight the importance of camera-based plant growth monitoring as (1) such systems are low-cost and (2) the collected data is crucial for nutrient management and harvesting scheduling; they develop a YOLOv8-based detection system for multi-plant basil growth tracking. In~\cite{ResNet50ButterheadLettureGrowthStage}, a ResNet-50-based butterhead lettuce growth stage classification system is presented.

Cucumber-specific research is an active area. \cite{CucumberRootAndSeendlingNeuralNetworks} investigate the relationship between cucumber root traits and crop yield to support the breeding of high-yielding, stress-tolerant cultivars. Cucumber production further benefits from automated harvesting systems. Accurate detection of harvest-ready cucumbers remains challenging due to the visual similarity between fruits and surrounding foliage, as well as frequent occlusions. To address this problem, \cite{YoloCucumberPlantDetection} propose a lightweight YOLO-based detection network tailored for cucumber harvesting scenarios. Quality assessment has also been explored using deep learning, with \cite{CucumberFruitClassificationCNN} introducing a CNN-based framework for classifying cucumber fruits into defective, curved, and straight/sound categories. For phenotyping applications, \cite{CucumberFruitKeypointForPhenotyping} present a lightweight keypoint detection model that enables automated extraction of fruit phenotypic traits. \cite{MyDistillationPlants} propose a pest classification method applicable to different plant species, including cucumber. Closest to our task, \cite{KhabarlakNalefChallenge2026} fine-tune a single DINOv2 ViT-L backbone with progressive layer unfreezing for cucumber growth stage classification.

In contrast to strongly supervised approaches, large language models (LLMs) have demonstrated remarkable progress through self-supervised learning, typically by training on large text corpora with a next-token prediction objective. This approach enables the network to learn strong linguistic representations from unlabeled data.

Vision Transformers have successfully adapted the concept of tokens to images, often surpassing convolutional neural networks. Initially, however, they were still trained with strong supervision, as the design of effective self-supervised objectives for vision is more challenging, and the notion of a ``next image token'' is less self-evident.

To pretrain a vision backbone with self-supervision, several pretext tasks have been proposed: image re-colorization, prediction of image transformations, and others. DINOv2 uses masked image modeling combined with a feature-consistency loss between teacher and student networks observing different crops of the same image. A large Vision Transformer (1.1B parameters~\cite{DINOv2}) is trained and then distilled into several smaller ViTs. This approach was later extended by DINOv3~\cite{DINOv3}, which identified limitations in DINOv2's architecture scaling under extended training and proposed a Gram loss, among other improvements. The resulting 7B model achieves state-of-the-art performance on many downstream tasks, and the authors distill it into both ViT and ConvNeXt variants. Models trained in this fashion are commonly referred to as Vision Foundation Models, as the rich features they produce transfer broadly across diverse downstream tasks.

The self-supervised training objective allows the collection of large unlabeled datasets for training. For instance, the authors of DINOv2 assembled the LVD-142M dataset comprising 142 million images without any labels, including ImageNet-22K and many other sources. Most of these are not plant-focused; while the authors include the small Flowers dataset, the larger nature-focused iNaturalist dataset is absent from the training set~\cite{DINOv2}. The DINOv3 authors introduce the LVD-1689M dataset (1,689 million Instagram photos). While substantially larger, this dataset likely entails a greater domain gap than DINOv2's web-sourced collection. Despite the domain gap, recent works have shown that self-supervised methods can be successfully adapted to plant-related tasks~\cite{GroundingDinoAgriculture,DinoV2AgricultureDomainShift}.

Despite these advances, most existing approaches to plant growth stage classification still rely on strongly supervised backbones, limiting their transferability to specialized agricultural domains with scarce labeled data. Furthermore, despite its agronomic importance, cucumber growth stage classification has received only limited attention in the literature, with VFM-based solutions for this specific task explored only in a single large-backbone configuration~\cite{KhabarlakNalefChallenge2026}. Finally, the task stands to benefit from additional in-model regularization, as classification must rely primarily on subtle leaf morphology cues due to frequent occlusion of flowers and fruits; the potential of lightweight ensemble head designs to improve the accuracy-efficiency trade-off in this setting remains unexplored.

\section{Materials and Methods}\label{sec:materials-methods}

To address plant growth stage classification, we develop an approach based on Vision Foundation Models paired with a regularizing multibranch ensemble head. We compare the proposed method against the architectures listed in \cref{tab:backbones}, spanning both convolutional neural networks (CNNs) and Transformer architectures.
\begin{table*}[htb]
  \caption{Backbones considered.}\label{tab:backbones}
  \centering
  \begin{tabular}{lll}
    \toprule
    Model                                & Type        & Sizes                          \\
    \midrule
    \multicolumn{3}{c}{Strongly Supervised Backbones}                                   \\
    \hline
    ResNet~\cite{ResNet}                 & CNN         & 50, 101                        \\
    ConvNeXt~\cite{ConvNeXt}             & CNN         & Tiny (T)                       \\
    EfficientNet~\cite{EfficientNet}     & CNN         & B3, B7                         \\
    EfficientNetV2~\cite{EfficientNetV2} & CNN         & Small (S), Large (L)           \\
    ViT~\cite{ViT}                       & Transformer & Base (B), Large (L)            \\
    \midrule
    \multicolumn{3}{c}{Self-Supervised Backbones}                                       \\
    \hline
    DINOv2 ViT~\cite{DINOv2}             & Transformer & Small (S), Large (L)           \\
    DINOv3 ViT~\cite{DINOv3}             & Transformer & Small (S), Large (L)           \\
    DINOv3 ConvNeXt~\cite{DINOv3}        & CNN         & Tiny (T), Small (S), Large (L) \\
    \bottomrule
  \end{tabular}
\end{table*}

\begin{figure}[tbh]
  \captionsetup[subfigure]{justification=centering}
  \centering
  \begin{subfigure}{0.19\linewidth}
    \centering
    \includegraphics[width=0.9\linewidth]{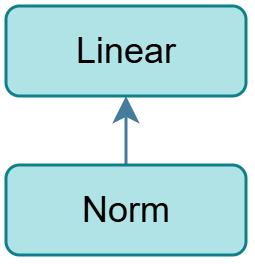}
    \caption{Light}\label{fig:heads-linear-probe}
  \end{subfigure}
  \begin{subfigure}{0.19\linewidth}
    \centering
    \includegraphics[width=0.9\linewidth]{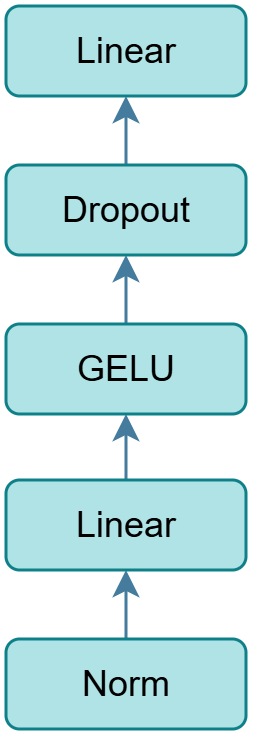}
    \caption{Heavy}\label{fig:heads-heavy}
  \end{subfigure}
  \begin{subfigure}{0.59\linewidth}
    \centering
    \includegraphics[width=0.9\linewidth]{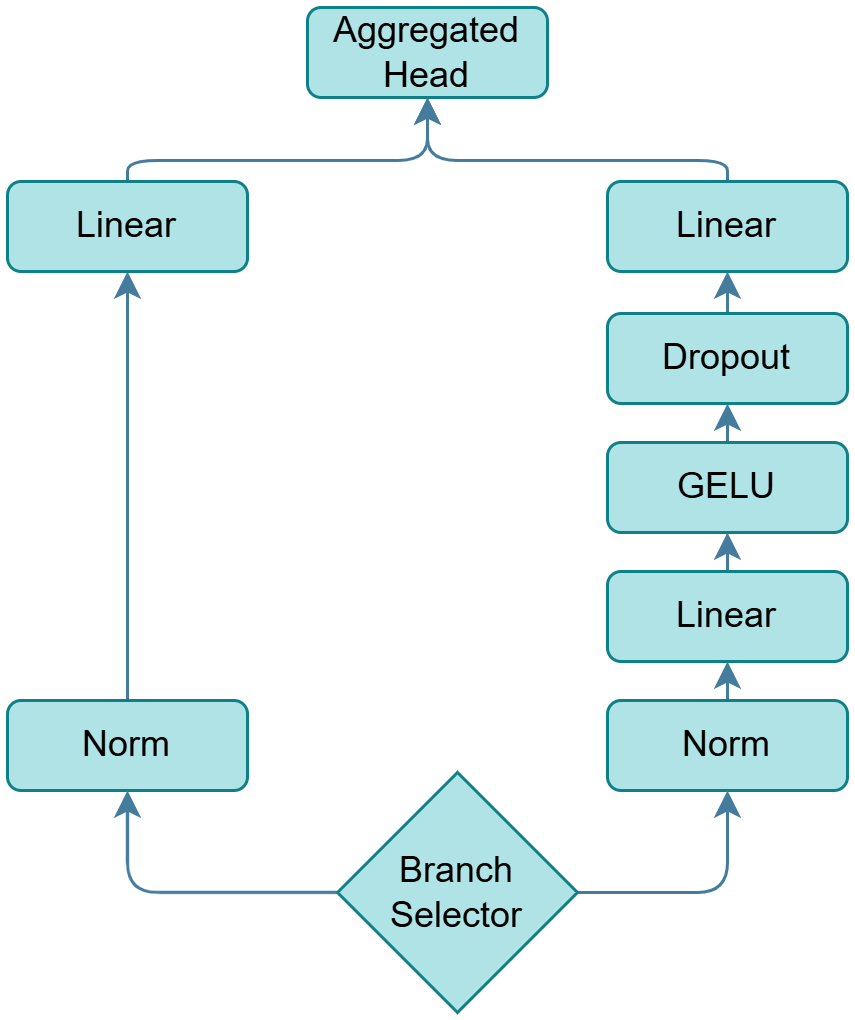}
    \caption{Multibranch Ensemble}\label{fig:heads-multibranch}
  \end{subfigure}
  \caption{Neural network heads proposed for fine-tuning.}\label{fig:heads}
\end{figure}

The classification head is an important component of the proposed architecture. Given the rich features extracted by the backbone, the head must adapt them to the downstream task. We develop three head variants. \cref{fig:heads-linear-probe} shows a lightweight head, also known as a linear probe, often used with VFMs as recommended in~\cite{DINOv2,DINOv3}, and frequently sufficient given the descriptive power of VFM features. To address the domain shift between VFM pretraining and downstream agricultural data, a head with a larger number of parameters may be required. \cref{fig:heads-heavy} is a heavy head comprising two MLP layers with GELU activation followed by Dropout regularization (probability 0.3). \cref{fig:heads-multibranch} shows the proposed multibranch ensemble head, described in detail below.

To improve the performance of VFMs, an ensemble of models can be used; however, this increases inference time by a factor of~\(N\). Several works instead build more robust representations within a single network, effectively forming an in-model ensemble. The prototypical example is Dropout~\cite{Dropout}: randomly masking activations during training samples a different weight-sharing subnetwork at each step, and test-time weight scaling approximates averaging the predictions of exponentially many such subnetworks. The authors of PTA~\cite{PTA} make this idea explicit, building an in-model ensemble from multiple parallel branches of different depths within the backbone, improving classification quality. StochasticDepth~\cite{StochasticDepth} demonstrates that dynamically varying network depth during training acts as a regularizer; this technique has been adopted by EfficientNet V1 and V2, DINOv2, and DINOv3. \cite{CollaborativeLearning} proposes collaborative networks where multiple branches are trained simultaneously, each supervised by a combination of ground-truth labels and soft targets from the other heads. \cite{MultiHeadMultiLoss} proposes a multi-head multi-loss classifier in which each head is assigned higher weight to a subset of classes, inducing specialization, with the final prediction obtained by averaging across heads.

In this work we propose a multibranch ensemble head (\cref{fig:heads-multibranch}). The two branches differ in depth and capacity: the left branch is a simple linear probe, while the right branch consists of several MLP layers with Dropout, equivalent to the heavy head described above (\cref{fig:heads-heavy}). During training, the active branch is randomly selected for each batch, so each branch learns to solve the task independently. The difference in capacity allows the light and heavy branches to capture simpler and more complex feature transformations, respectively. At inference, an in-model ensemble is formed by aggregating the outputs of both branches:

\begin{equation}
  \label{eq:ensemble-head-averaging}
  \hat{y} = \log \left(\frac{1}{2} \left(p^{(\mathrm{light})} + p^{(\mathrm{heavy})} \right) + \epsilon\right),
\end{equation}
where the branch output \(p^{(k)},\ k \in \{\mathrm{light}, \mathrm{heavy}\}\) is defined as:
\begin{equation}
  \label{eq:head-output}
  p^{(k)} = \mathrm{softmax} \left( f^{(k)}(z) \right).
\end{equation}
Here \(f^{(k)}\) is the function computed by branch \(k\), and \(z = g(X)\) are the features produced by the backbone \(g\) for the input batch \(X\). The vector \(\hat{y} \in \mathbb{R}^{C}\) holds the aggregated log-probabilities, and \(\epsilon\) is a small positive constant that ensures numerical stability of the logarithm. Aggregation is performed in probability space rather than logit space (Eqs.~\ref{eq:ensemble-head-averaging}--\ref{eq:head-output}), since logits from branches of different capacity may have substantially different scales. Probability-space aggregation is consistent with standard mixture-of-experts averaging and corresponds to a uniform Bayesian model average treating each head as an equally plausible hypothesis about the class distribution.

The branch selector shown in \cref{fig:heads-multibranch} activates exactly one branch per training step. At each step a branch is drawn independently from a Bernoulli distribution, \(k \sim \mathrm{Bernoulli}(\pi)\), with \(\pi = 0.5\) assigning equal probability to the two branches. Only the selected branch produces the prediction for that step, and only its parameters, together with those of the backbone, receive gradients from the cross-entropy loss; the other branch is left unchanged. No knowledge-distillation or cross-branch consistency term is applied, so each branch learns to solve the task independently. \cref{alg:multibranch} summarizes the procedure.

\begin{algorithm}[tb]
  \DontPrintSemicolon
  \caption{Training step with the multibranch ensemble head}\label{alg:multibranch}
  \Input{mini-batch \((X, Y)\); backbone \(g\); branches \(f^{(\mathrm{light})}, f^{(\mathrm{heavy})}\); branch probability \(\pi\)}
  \(z \leftarrow g(X)\) \tcp*{backbone features}
  draw \(k \sim \mathrm{Bernoulli}(\pi)\), \quad \(k \in \{\mathrm{light}, \mathrm{heavy}\}\)\;
  \(\ell \leftarrow \mathrm{CrossEntropy}\!\left(f^{(k)}(z),\, Y\right)\)\;
  update parameters of \(g\) and \(f^{(k)}\) using \(\nabla \ell\) \tcp*{other branch weights are not updated}
\end{algorithm}

Models are trained using the AdamW optimizer with a batch size of 32. The head is randomly initialized while the backbone uses pretrained weights. To mitigate catastrophic forgetting, training proceeds in two stages: 15 epochs of head-only fine-tuning at a learning rate of $3 \times 10^{-3}$, followed by 15 epochs of full-network fine-tuning at a reduced learning rate of $5 \times 10^{-6}$.

\section{Experiments}

To evaluate the proposed approach we use the NALEF Food Safety dataset~\cite{NalefPlantGrowth}, published in 2026. The dataset contains images of cucumber crops captured under real-world conditions, annotated with five growth stages: plantula, early vegetative, vegetative growth, flowering, and fruit setting. Overall there are 787 training and 239 test images. As the public test set does not include ground-truth annotations, we perform a stratified train/validation split with an 80/20 ratio for all reported experiments. Images are captured using different devices (professional and smartphone cameras) under varying environmental conditions.

The dataset is strongly imbalanced across growth stages, as summarized in \cref{tab:class-distribution}. The smallest class (plantula) contributes an order of magnitude fewer samples than the largest (vegetative growth), leaving only about ten validation images for the minority stages. Accuracy is unsuitable as the primary metric for highly imbalanced datasets. We therefore evaluate all models using Macro F1 score and Cohen's Kappa.

\begin{table}[htb]
  \caption{Per-class image counts for the stratified 80/20 train/validation split.}\label{tab:class-distribution}
  \centering
  \begin{tabular}{lrrr}
    \toprule
    Growth stage        & Train & Val & Total \\
    \midrule
    Plantula            &  40   &  10 &  50   \\
    Early vegetative    &  42   &  11 &  53   \\
    Vegetative growth   & 307   &  77 & 384   \\
    Flowering           &  70   &  17 &  87   \\
    Fruit setting       & 170   &  43 & 213   \\
    \midrule
    Total               & 629   & 158  & 787  \\
    \bottomrule
  \end{tabular}
\end{table}

To improve model generalization, the following augmentations are applied: randomly resized crop, random horizontal flip, and color jitter. All images are resized to \(224 \times 224\).

\section{Results}

\begin{figure}[tbh]
  \centering
  \includegraphics[width=\linewidth]{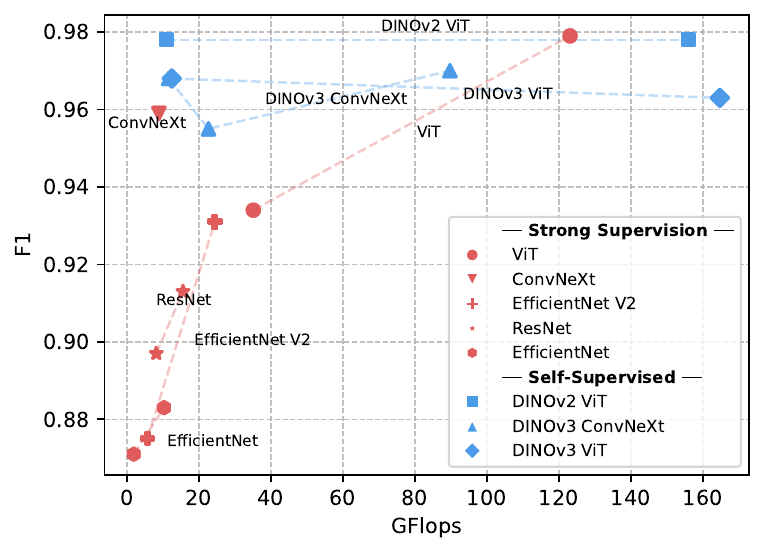}
  \caption{Macro F1 score vs.\ model computational complexity (GFlops). Self-supervised backbones (blue) hold a clear advantage in the small- and mid-compute regime.}\label{fig:results-plot}
\end{figure}

Experiment results are presented in \cref{tab:results}.
\begin{table*}[tbh]
  \caption{Evaluation results of the considered models and heads. DINOv2 ViT-S with the multibranch head (highlighted in red) achieves the best quality-efficiency trade-off.}\label{tab:results}
  \centering
  \begin{tabular}{lllrrrr}
    \toprule
    Model                 & Head          & B. Supervision & F1    & Kappa & GFlops & Params (M) \\
    \midrule
    ViT-L                 & Heavy         & Strong         & 0.979 & 0.972 & 123.108   & 303.831 \\
    DINOv2 ViT-L          & Heavy         & Self           & 0.978 & 0.971 & 162.025   & 304.898 \\
    \best{DINOv2 ViT-S}   & Multibranch   & Self           & 0.978 & 0.971 &  12.248   &  22.260 \\
    DINOv3 ConvNeXt-L     & Light         & Self           & 0.970 & 0.962 &  68.720   & 196.238 \\
    DINOv3 ViT-S          & Heavy         & Self           & 0.968 & 0.953 &  9.397   &  21.797 \\
    DINOv3 ConvNeXt-T     & Light         & Self           & 0.968 & 0.962 &  8.910   &  27.824 \\
    DINOv3 ViT-L          & Heavy         & Self           & 0.963 & 0.952 & 125.681   & 303.659 \\
    ConvNeXt-T            & Heavy         & Strong         & 0.959 & 0.952 &   8.910   &  28.218 \\
    DINOv3 ConvNeXt-S     & Light         & Self           & 0.955 & 0.943 &  17.366   &  49.459 \\
    ViT-B                 & Heavy         & Strong         & 0.934 & 0.914 &  35.127   &  86.196 \\
    EfficientNet V2-L     & Heavy         & Strong         & 0.931 & 0.894 &  24.461   & 117.893   \\
    ResNet101             & Multibranch   & Strong         & 0.913 & 0.875 &  15.601   &  43.562   \\
    ResNet50              & Heavy         & Strong         & 0.897 & 0.874 &   8.176   &  24.560   \\
    EfficientNet-B7       & Heavy         & Strong         & 0.883 & 0.874 &  10.337   &  65.101   \\
    EfficientNet V2-S     & Heavy         & Strong         & 0.875 & 0.845 &   5.699   &  20.836   \\
    EfficientNet-B3       & Light         & Strong         & 0.871 & 0.810 &   1.922   &  10.704   \\
    \bottomrule
  \end{tabular}
\end{table*}
For each model we report the head configuration that achieved the best result, the type of backbone supervision, evaluation metrics, and computational complexity. For strongly supervised models we report the model name and size; for self-supervised backbones we report the DINO version, backbone architecture, and size. Most strongly supervised backbones are outperformed by self-supervised ones of comparable size. The exception is ViT-L, which is on par with the best DINO-pretrained models and attains the highest Macro F1 overall (0.979). Therefore, the advantage of self-supervised pretraining is concentrated in the small- and mid-sized models (\cref{fig:results-plot}).

\cref{fig:results-plot} shows Macro F1 score as a function of computational complexity (GFlops).
Strongly supervised models are shown in red and self-supervised models in blue. The advantage of DINO-pretrained backbones is most pronounced in the small-model regime. The best result for a small model (Macro F1 of 0.978) is achieved by DINOv2 ViT-S with the proposed multibranch ensemble head (highlighted in red in \cref{tab:results}), matching the strongly supervised ViT-L model, which is over \(10\times\) larger. A similarly sized ConvNeXt-Tiny model achieves a substantially lower Macro F1 of 0.959; DINOv3-style pretraining of ConvNeXt-Tiny raises this to 0.968.

DINOv2 models also outperform their DINOv3 ViT and ConvNeXt counterparts. As discussed above, DINOv3 models are trained on a larger dataset sourced primarily from Instagram, and the greater domain shift is the likely cause of their slightly inferior performance.

\section{Discussion}

\cref{tab:ablation} details the incremental improvements to DINOv2 ViT-S proposed in this work. As a baseline we train DINOv2 ViT-S with a light head (\cref{fig:heads-linear-probe}) for 15 epochs.
\begin{table*}[htb]
  \caption{Ablation study of the proposed improvements over DINOv2 ViT-S.}\label{tab:ablation}
  \centering
  \begin{tabular}{lrrrrr}
    \toprule
    Modification            & F1    & Kappa & GFlops & Params (M)  \\
    \midrule
    DINOv2 ViT-S (baseline) & 0.915 & 0.895 & 12.247 & 22.059      \\
    + multibranch head      & 0.960 & 0.944 & 12.248 & 22.260      \\
    + backbone fine-tuning  & 0.978 & 0.971 & 12.248 & 22.260      \\
    \bottomrule
  \end{tabular}
\end{table*}
Replacing the linear probe with a heavy head improves F1 to 0.955. Switching to the multibranch head further raises F1 to 0.960. Subsequently fine-tuning the full backbone yields the final F1 of 0.978. Importantly, the multibranch head adds negligible parameters and essentially no additional inference overhead compared to the single-branch heavy head, since backbone computation dominates.

\begin{figure}[tbh]
  \centering
  \includegraphics[width=\linewidth]{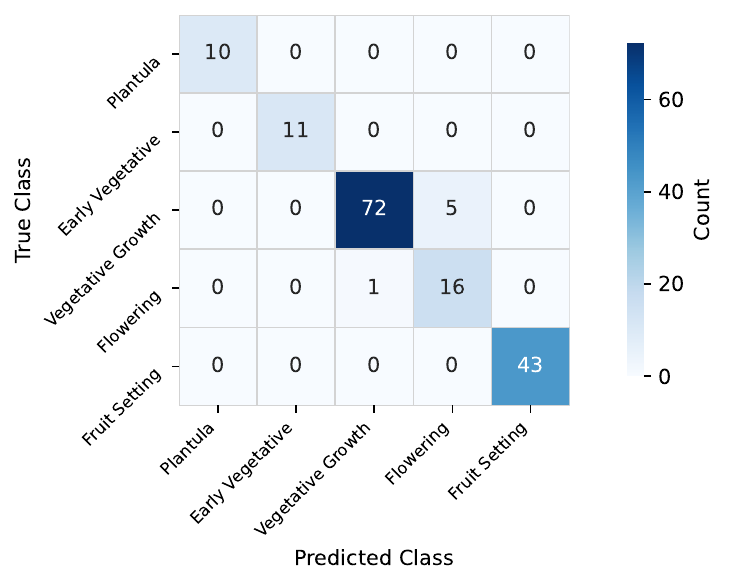}
  \caption{Confusion matrix of DINOv2 ViT-S with multibranch head.}\label{fig:confusion-matrix}
\end{figure}

As shown in \cref{fig:confusion-matrix}, the proposed DINOv2 ViT-S with multibranch head is a robust classifier: it makes no errors between non-adjacent growth stages, and despite significant class imbalance it remains accurate on minority classes.

The only misclassifications occur between the vegetative growth and flowering stages. At first glance these stages should be visually distinct, separable by the presence of flowers. However, closer inspection of the NALEF Food Safety dataset reveals that images primarily show cucumber leaves, with flowers or fruits often occluded by foliage. The model must therefore discriminate stages based on subtle differences in leaf and vine appearance alone, which is a considerably harder task.

\paragraph{Limitations.} The dataset is captured from a leaf-centric viewpoint in which flowers and fruits are frequently occluded, which is also the main source of the residual vegetative-growth vs. flowering confusion. Typically, field examination is performed during daytime and under good environmental conditions. We have not evaluated robustness to adverse conditions such as heavy shading, rain, night-time capture, or lens contamination on smartphone and edge devices. In this work we target the NALEF Food Safety dataset. Therefore, retraining will be needed to adapt the model to other cultivars or environmental conditions. Also, the validation split is highly imbalanced, so small differences in Macro F1 between the top models should be interpreted with this in mind.

The multibranch ensemble head targets any fine-grained classification task where a lightweight head on top of frozen or fine-tuned features must be trained on limited data and is at risk of overfitting. Applying and evaluating it on other fine-grained benchmarks, for instance, species recognition or disease severity grading, is a natural next step that would establish how far the regularization benefit observed here carries over.

\section{Conclusions}

In this work we proposed a regularizing multibranch ensemble head for adapting Vision Foundation Models to small, domain-shifted datasets, and studied it on cucumber growth stage classification as a fine-grained case study. Experiments across a broad set of strongly supervised and self-supervised backbones on the NALEF Food Safety dataset indicate that self-supervised pretraining yields stronger feature representations for this task in the small- and mid-sized model regime. DINOv2 and DINOv3 backbones outperform ResNet, EfficientNet, and strongly supervised ViT variants of comparable size, while at the largest scale the strongly supervised ViT-L remains competitive; overall, VFMs transfer well to agricultural imagery despite substantial pre-training domain gaps, with the clearest benefit when model capacity is limited. Notably, DINOv2 ViT-S paired with the proposed multibranch ensemble head achieves a Macro F1 of 0.978 and Cohen's Kappa of 0.971, matching models with over 300M parameters while requiring only 12.2~GFlops and 22.3M parameters, making it suitable for deployment on smartphones and edge devices.

The introduced multibranch ensemble head trains a lightweight linear branch and a heavier MLP branch simultaneously via stochastic branch sampling.  At inference, predictions are aggregated in probability space as a uniform Bayesian model average. Ablation analysis confirms that the multibranch head improves over both the linear probe and the heavy head, and that subsequent full-network fine-tuning yields further substantial gains. The method adds negligible parameters and essentially zero inference overhead compared to a single-branch heavy head.

Error analysis reveals that misclassifications are confined to adjacent phenological stages, specifically between vegetative growth and flowering, a consequence of the dataset's leaf-centric imaging perspective in which flowers and fruits are frequently occluded. The absence of errors between non-adjacent stages indicates well-structured internal representations.

Future work includes extending the approach to additional crop species and growth stage taxonomies, as well as incorporating explainability by integrating large language models.

\begin{small}
  \subsection*{Funding}

  This research was carried out as part of the scientific project ``Software and algorithmic solutions for agricultural information and analytical systems based on artificial intelligence in variable agroclimatic conditions'' (state registration number 0126U001122) funded by the Ministry of Education and Science of Ukraine at the expense of the state budget.

  \subsection*{Declaration on Generative AI}

  During the preparation of this work, the author used ChatGPT-5.6 for grammar and spelling checks. The author subsequently reviewed and edited the content as necessary and takes full responsibility for the content of this publication.

  \printbibliography

@article{BasilYoloGrowthStage,
  title        = {Instance-Level Phenotype-Based Growth Stage Classification of Basil in Multi-Plant Environments},
  author       = {Kim, Jung-Sun Gloria and Shin, Soo Hyun and Chung, Soo},
  date         = {2025-11-18},
  journaltitle = {Frontiers in Plant Science},
  shortjournal = {Front. Plant Sci.},
  volume       = {16},
  pages        = {1707985},
  issn         = {1664-462X},
  doi          = {10.3389/fpls.2025.1707985},
  url          = {https://www.frontiersin.org/articles/10.3389/fpls.2025.1707985/full}
}

@inproceedings{CollaborativeLearning,
  author    = {Song, Guocong and Chai, Wei},
  title     = {Collaborative learning for deep neural networks},
  year      = {2018},
  publisher = {Curran Associates Inc.},
  address   = {Red Hook, NY, USA},
  booktitle = {Proceedings of the 32nd International Conference on Neural Information Processing Systems},
  pages     = {1837–1846},
  numpages  = {10},
  location  = {Montr\'{e}al, Canada},
  series    = {NIPS'18}
}

@inproceedings{ConvNeXt,
  title     = {A {{ConvNet}} for the 2020s},
  booktitle = {{{IEEE}}/{{CVF}} Conference on Computer Vision and Pattern Recognition, {{CVPR}} 2022, New Orleans, {{LA}}, {{USA}}, June 18-24, 2022},
  author    = {Liu, Zhuang and Mao, Hanzi and Wu, Chao-Yuan and Feichtenhofer, Christoph and Darrell, Trevor and Xie, Saining},
  date      = {2022},
  pages     = {11966--11976},
  publisher = {IEEE},
  doi       = {10.1109/CVPR52688.2022.01167},
  url       = {https://doi.org/10.1109/CVPR52688.2022.01167}
}

@article{CucumberDiseaseDetection,
  title        = {Deep Learning Method for Cucumber Disease Detection in Complex Environments for New Agricultural Productivity},
  author       = {Liu, Jun and Wang, Xuewei and Chen, Qian and Yan, Peng and Liu, Xin},
  date         = {2025-07-07},
  journaltitle = {BMC Plant Biology},
  shortjournal = {BMC Plant Biol},
  volume       = {25},
  number       = {1},
  pages        = {888},
  issn         = {1471-2229},
  doi          = {10.1186/s12870-025-06841-y},
  url          = {https://bmcplantbiol.biomedcentral.com/articles/10.1186/s12870-025-06841-y},
  urldate      = {2026-05-25},
  langid       = {english}
}

@article{CucumberFruitClassificationCNN,
  title        = {A {{New Method}} to {{Optimize Deep CNN Model}} for {{Classification}} of {{Regular Cucumber Based}} on {{Global Average Pooling}}},
  author       = {Haseli Golzar, Sajad and Bagherpour, Hossein and Amiri Parian, Jafar},
  date         = {2024-01},
  journaltitle = {Journal of Food Processing and Preservation},
  shortjournal = {Journal of Food Processing and Preservation},
  volume       = {2024},
  number       = {1},
  pages        = {5818803},
  issn         = {0145-8892, 1745-4549},
  doi          = {10.1155/2024/5818803},
  url          = {https://onlinelibrary.wiley.com/doi/10.1155/2024/5818803},
  urldate      = {2026-05-25},
  langid       = {english}
}

@article{CucumberFruitKeypointForPhenotyping,
  title        = {{{YOLO11m-SCFPose}}: {{An Improved Detection Framework}} for {{Keypoint Extraction}} in {{Cucumber Fruit Phenotyping}}},
  shorttitle   = {{{YOLO11m-SCFPose}}},
  author       = {Yu, Huijiao and Zhang, Xuehui and Yan, Jun and Meng, Xianyong},
  date         = {2025-07-20},
  journaltitle = {Horticulturae},
  shortjournal = {Horticulturae},
  volume       = {11},
  number       = {7},
  pages        = {858},
  issn         = {2311-7524},
  doi          = {10.3390/horticulturae11070858},
  url          = {https://www.mdpi.com/2311-7524/11/7/858},
  urldate      = {2026-05-25},
  langid       = {english}
}

@article{CucumberRootAndSeendlingNeuralNetworks,
  title        = {Deep Learning‐based Association Analysis of Root Image Data and Cucumber Yield},
  author       = {Zhu, Cuifang and Yu, Hongjun and Lu, Tao and Li, Yang and Jiang, Weijie and Li, Qiang},
  date         = {2024-05},
  journaltitle = {The Plant Journal},
  shortjournal = {The Plant Journal},
  volume       = {118},
  number       = {3},
  pages        = {696--716},
  issn         = {0960-7412, 1365-313X},
  doi          = {10.1111/tpj.16627},
  url          = {https://onlinelibrary.wiley.com/doi/10.1111/tpj.16627},
  urldate      = {2026-05-25},
  langid       = {english}
}

@article{DINOv2,
  author    = {Maxime Oquab and
               Timoth{\'{e}}e Darcet and
               Th{\'{e}}o Moutakanni and
               Huy V. Vo and
               Marc Szafraniec and
               Vasil Khalidov and
               Pierre Fernandez and
               Daniel Haziza and
               Francisco Massa and
               Alaaeldin El{-}Nouby and
               Mido Assran and
               Nicolas Ballas and
               Wojciech Galuba and
               Russell Howes and
               Po{-}Yao Huang and
               Shang{-}Wen Li and
               Ishan Misra and
               Michael Rabbat and
               Vasu Sharma and
               Gabriel Synnaeve and
               Hu Xu and
               Herv{\'{e}} J{\'{e}}gou and
               Julien Mairal and
               Patrick Labatut and
               Armand Joulin and
               Piotr Bojanowski},
  title     = {DINOv2: Learning Robust Visual Features without Supervision},
  journal   = {Trans. Mach. Learn. Res.},
  volume    = {2024},
  year      = {2024},
  url       = {https://openreview.net/forum?id=a68SUt6zFt},
  bibsource = {dblp computer science bibliography, https://dblp.org}
}

@misc{DinoV2AgricultureDomainShift,
  title         = {Mitigating Domain Drift in Multi Species Segmentation with DINOv2: A Cross-Domain Evaluation in Herbicide Research Trials},
  author        = {Artzai Picon and Itziar Eguskiza and Daniel Mugica and Javier Romero and Carlos Javier Jimenez and Eric White and Gabriel Do-Lago-Junqueira and Christian Klukas and Ramon Navarra-Mestre},
  year          = {2026},
  eprint        = {2508.07514},
  archiveprefix = {arXiv},
  primaryclass  = {cs.CV},
  url           = {https://arxiv.org/abs/2508.07514}
}

@article{DINOv3,
  author     = {Oriane Sim{\'{e}}oni and
                Huy V. Vo and
                Maximilian Seitzer and
                Federico Baldassarre and
                Maxime Oquab and
                Cijo Jose and
                Vasil Khalidov and
                Marc Szafraniec and
                Seung Eun Yi and
                Micha{\"{e}}l Ramamonjisoa and
                Francisco Massa and
                Daniel Haziza and
                Luca Wehrstedt and
                Jianyuan Wang and
                Timoth{\'{e}}e Darcet and
                Th{\'{e}}o Moutakanni and
                Leonel Sentana and
                Claire Roberts and
                Andrea Vedaldi and
                Jamie Tolan and
                John Brandt and
                Camille Couprie and
                Julien Mairal and
                Herv{\'{e}} J{\'{e}}gou and
                Patrick Labatut and
                Piotr Bojanowski},
  title      = {DINOv3},
  journal    = {CoRR},
  volume     = {abs/2508.10104},
  year       = {2025},
  url        = {https://doi.org/10.48550/arXiv.2508.10104},
  doi        = {10.48550/ARXIV.2508.10104},
  eprinttype = {arXiv},
  eprint     = {2508.10104},
  bibsource  = {dblp computer science bibliography, https://dblp.org}
}

@article{Dropout,
  title        = {Improving Neural Networks by Preventing Co-Adaptation of Feature Detectors},
  author       = {Hinton, Geoffrey E. and Srivastava, Nitish and Krizhevsky, Alex and Sutskever, Ilya and Salakhutdinov, Ruslan},
  date         = {2012},
  journaltitle = {CoRR},
  volume       = {abs/1207.0580},
  eprint       = {1207.0580},
  eprinttype   = {arXiv},
  url          = {http://arxiv.org/abs/1207.0580},
  bibsource    = {dblp computer science bibliography, https://dblp.org}
}

@inproceedings{EfficientNet,
  title     = {{{EfficientNet}}: {{Rethinking}} Model Scaling for Convolutional Neural Networks},
  booktitle = {Proceedings of the 36th {{International Conference}} on {{Machine Learning}}, {{ICML}} 2019, 9-15 {{June}} 2019, {{Long Beach}}, {{California}}, {{USA}}},
  author    = {Tan, Mingxing and Le, Quoc V.},
  date      = {2019},
  series    = {Proceedings of Machine Learning Research},
  volume    = {97},
  pages     = {6105--6114},
  publisher = {PMLR},
  bibsource = {dblp computer science bibliography, https://dblp.org}
}

@article{EfficientNetStrawBerryGrowthClassification,
  title        = {{{SmartBerry}} for {{AI-based}} Growth Stage Classification and Precision Nutrition Management in Strawberry Cultivation},
  author       = {Darlan, Daison and Ajani, Oladayo S. and An, Joon Woo and Bae, Nan Yeon and Lee, Bram and Park, Tusan and Mallipeddi, Rammohan},
  date         = {2025-04-23},
  journaltitle = {Scientific Reports},
  shortjournal = {Sci Rep},
  volume       = {15},
  number       = {1},
  pages        = {14019},
  issn         = {2045-2322},
  doi          = {10.1038/s41598-025-97168-z},
  url          = {https://www.nature.com/articles/s41598-025-97168-z},
  langid       = {english}
}

@inproceedings{EfficientNetV2,
  title     = {{{EfficientNetV2}}: {{Smaller}} Models and Faster Training},
  booktitle = {Proceedings of the 38th {{International Conference}} on {{Machine Learning}}, {{ICML}} 2021, 18-24 {{July}} 2021, {{Virtual Event}}},
  author    = {Tan, Mingxing and Le, Quoc V.},
  date      = {2021},
  volume    = {139},
  pages     = {10096--10106},
  publisher = {PMLR},
  bibsource = {dblp computer science bibliography, https://dblp.org}
}

@inproceedings{GroundingDinoAgriculture,
  title     = {Few-Shot Adaptation of Grounding {{DINO}} for Agricultural Domain},
  booktitle = {{{IEEE}}/{{CVF}} Conference on Computer Vision and Pattern Recognition Workshops, {{CVPR}} Workshops 2025, Nashville, {{TN}}, {{USA}}, June 11-15, 2025},
  author    = {Singh, Rajhans and Bidese-Puhl, Rafael and Dhakal, Kshitiz and Sornapudi, Sudhir},
  date      = {2025},
  pages     = {5332--5342},
  publisher = {Computer Vision Foundation / IEEE},
  url       = {https://openaccess.thecvf.com/content/CVPR2025W/V4A/html/Singh_Few-Shot_Adaptation_of_Grounding_DINO_for_Agricultural_Domain_CVPRW_2025_paper.html},
  bibsource = {dblp computer science bibliography, https://dblp.org}
}

@inproceedings{MultiHeadMultiLoss,
  author    = {Galdran, Adrian
               and Verjans, Johan W.
               and Carneiro, Gustavo
               and Gonz{\'a}lez Ballester, Miguel A.},
  title     = {Multi-Head Multi-Loss Model Calibration},
  booktitle = {Medical Image Computing and Computer Assisted Intervention -- MICCAI 2023},
  year      = {2023},
  publisher = {Springer Nature Switzerland},
  address   = {Cham},
  pages     = {108--117},
  isbn      = {978-3-031-43898-1}
}

@inproceedings{MyDistillationPlants,
  title     = {Feature Knowledge Distillation Using Group Convolutions for Efficient Plant Pest Recognition},
  booktitle = {Proceedings of the 12th {{International Conference Information Control Systems}} \& {{Technologies}} ({{ICST}} 2024), {{Odesa}}, {{Ukraine}}, {{September}} 23–25},
  author    = {Khabarlak, Kostiantyn and Laktionov, Ivan and Diachenko, Grygorii},
  date      = {2024},
  series    = {{{CEUR}} Workshop Proceedings},
  volume    = {3790},
  pages     = {377--387},
  url       = {https://ceur-ws.org/Vol-3790/paper33.pdf}
}

@dataset{NalefPlantGrowth,
  author    = {Menco Tovar, Andrea Carolina and
               Puertas, Edwin and
               Martinez-Santos, Juan Carlos and
               Serrano Castañeda, Jairo},
  title     = {NALEF Food Safety 2026},
  month     = mar,
  year      = 2026,
  publisher = {Zenodo},
  doi       = {10.5281/zenodo.19192941},
  url       = {https://doi.org/10.5281/zenodo.19192941}
}

@inproceedings{KhabarlakNalefChallenge2026,
  title     = {Fine-Tuned {DINOv2} {ViT-L} for Cucumber Growth Stage Classification: {NALEF} Food Safety Challenge Solution},
  booktitle = {Proceedings of the {NALEF} 2026 Workshop, Cartagena, Colombia},
  author    = {Khabarlak, Kostiantyn},
  year      = 2026,
  series    = {CEUR Workshop Proceedings},
  note      = {In press}
}

@article{PlantPhenologicalResearch2022,
  author  = {Katal, Negin  and Rzanny, Michael  and Mäder, Patrick  and Wäldchen, Jana },
  title   = {Deep Learning in Plant Phenological Research: A Systematic Literature Review},
  journal = {Frontiers in Plant Science},
  volume  = {Volume 13 - 2022},
  year    = {2022},
  url     = {https://www.frontiersin.org/journals/plant-science/articles/10.3389/fpls.2022.805738},
  doi     = {10.3389/fpls.2022.805738},
  issn    = {1664-462X}
}

@inproceedings{PTA,
  title     = {Post-Train Adaptive {{MobileNet}} for Fast Anti-Spoofing},
  booktitle = {Proceedings of the 3rd {{International Workshop}} on {{Intelligent Information Technologies}} \& {{Systems}} of {{Information Security}}, {{Khmelnytskyi}}, {{Ukraine}}, {{March}} 23-25, 2022},
  author    = {Khabarlak, Kostiantyn},
  date      = {2022},
  series    = {{{CEUR}} Workshop Proceedings},
  volume    = {3156},
  pages     = {44--53},
  publisher = {CEUR-WS.org},
  url       = {http://ceur-ws.org/Vol-3156/keynote5.pdf},
  bibsource = {dblp computer science bibliography, https://dblp.org}
}

@inproceedings{ResNet,
  title     = {Deep Residual Learning for Image Recognition},
  booktitle = {2016 {{IEEE Conference}} on {{Computer Vision}} and {{Pattern Recognition}}, {{CVPR}} 2016, {{Las Vegas}}, {{NV}}, {{USA}}, {{June}} 27-30, 2016},
  author    = {He, Kaiming and Zhang, Xiangyu and Ren, Shaoqing and Sun, Jian},
  date      = {2016},
  pages     = {770--778},
  publisher = {IEEE Computer Society},
  doi       = {10.1109/CVPR.2016.90},
  bibsource = {dblp computer science bibliography, https://dblp.org}
}

@article{ResNet50ButterheadLettureGrowthStage,
  title        = {Comparison of {{Image Preprocessing Strategies}} for {{Convolutional Neural Network-Based Growth Stage Classification}} of {{Butterhead Lettuce}} in {{Industrial Plant Factories}}},
  author       = {Kim, Jung-Sun Gloria and Chung, Soo and Ko, Myungjin and Song, Jihoon and Shin, Soo Hyun},
  date         = {2025-06-03},
  journaltitle = {Applied Sciences},
  shortjournal = {Applied Sciences},
  volume       = {15},
  number       = {11},
  pages        = {6278},
  issn         = {2076-3417},
  doi          = {10.3390/app15116278},
  url          = {https://www.mdpi.com/2076-3417/15/11/6278}
}

@online{Sentinel2,
  title   = {Sentinel-2 | {{Copernicus Data Space Ecosystem}}},
  date    = {2026-05-14},
  url     = {https://dataspace.copernicus.eu/data-collections/copernicus-sentinel-missions/sentinel-2},
  urldate = {2026-05-14}
}

@inproceedings{StochasticDepth,
  title     = {Deep Networks with Stochastic Depth},
  booktitle = {Computer {{Vision}} - {{ECCV}} 2016 - 14th {{European Conference}}, {{Amsterdam}}, the {{Netherlands}}, {{October}} 11-14, 2016, {{Proceedings}}, {{Part IV}}},
  author    = {Huang, Gao and Sun, Yu and Liu, Zhuang and Sedra, Daniel and Weinberger, Kilian Q.},
  date      = {2016},
  series    = {Lecture Notes in Computer Science},
  volume    = {9908},
  pages     = {646--661},
  publisher = {Springer},
  doi       = {10.1007/978-3-319-46493-0\_39},
  bibsource = {dblp computer science bibliography, https://dblp.org}
}

@inproceedings{ViT,
  author    = {Alexey Dosovitskiy and
               Lucas Beyer and
               Alexander Kolesnikov and
               Dirk Weissenborn and
               Xiaohua Zhai and
               Thomas Unterthiner and
               Mostafa Dehghani and
               Matthias Minderer and
               Georg Heigold and
               Sylvain Gelly and
               Jakob Uszkoreit and
               Neil Houlsby},
  title     = {An Image is Worth 16x16 Words: Transformers for Image Recognition
               at Scale},
  booktitle = {9th International Conference on Learning Representations, {ICLR} 2021,
               Virtual Event, Austria, May 3-7, 2021},
  publisher = {OpenReview.net},
  year      = {2021},
  url       = {https://openreview.net/forum?id=YicbFdNTTy},
  bibsource = {dblp computer science bibliography, https://dblp.org}
}

@article{YoloCucumberPlantDetection,
  title        = {{{LMS-Res-YOLO}}: {{Lightweight}} and {{Multi-Scale Cucumber Detection Model}} with {{Residual Blocks}}},
  shorttitle   = {{{LMS-Res-YOLO}}},
  author       = {Li, Bo and Zhong, Guangjin and Ke, Wei},
  date         = {2025-12-01},
  journaltitle = {Sensors},
  shortjournal = {Sensors},
  volume       = {25},
  number       = {23},
  pages        = {7305},
  issn         = {1424-8220},
  doi          = {10.3390/s25237305},
  url          = {https://www.mdpi.com/1424-8220/25/23/7305},
  urldate      = {2026-05-25},
  langid       = {english}
}

\end{small}

\end{document}